\documentclass{article}

\usepackage{arxiv}

\usepackage[utf8]{inputenc} 
\usepackage[T1]{fontenc}    
\usepackage{hyperref}       
\usepackage{url}            
\usepackage{booktabs}       
\usepackage{amsfonts}       
\usepackage{nicefrac}       
\usepackage{microtype}      
\usepackage{lipsum}		
\usepackage{graphicx}
\usepackage{natbib}
\usepackage{doi}
\usepackage{pifont}

\title{Experiments in Artificial Culture: from noisy imitation to storytelling robots
}

\author{ \href{https://orcid.org/0000-0002-1476-3127}{\includegraphics[scale=0.06]{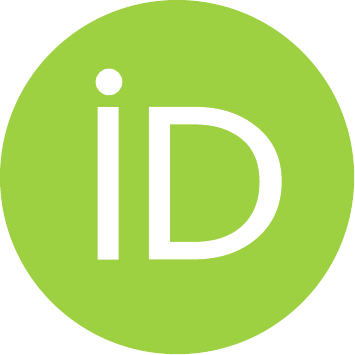}\hspace{1mm}Alan F.T. Winfield} \\
	Bristol Robotics Laboratory\\
	University of the West of England\\
	Bristol BS16 1QY \\
	\texttt{alan.winfield@brl.ac.uk} \\
	\And
	\href{https://orcid.org/0000-0003-0845-2223}{\includegraphics[scale=0.06]{orcid.pdf}\hspace{1mm}Susan Blackmore} \\
	Department of Psychology\\
	University of Plymouth\\
	Plymouth PL4 8AA \\ \\
}

\renewcommand{\shorttitle}{Experiments in Artificial Culture}

\hypersetup{
pdftitle={A template for the arxiv style},
pdfsubject={q-bio.NC, q-bio.QM},
pdfauthor={David S.~Hippocampus, Elias D.~Striatum},
pdfkeywords={First keyword, Second keyword, More},
}

\begin{document}
\maketitle

\begin{abstract}
	This paper presents a series of experiments in collective social robotics, spanning more than 10 years, with the long-term aim of building embodied models of (aspects) of cultural evolution. Initial experiments demonstrated the emergence of behavioural traditions in a group of social robots programmed to imitate each other’s behaviours (we call these Copybots). These experiments show that the noisy (i.e. less than perfect fidelity) imitation that comes for free with real physical robots gives rise naturally to variation in social learning. More recent experimental work extends the robots’ cognitive capabilities with simulation-based internal models, equipping them with a simple artificial theory of mind. With this extended capability we explore, in our current work, social learning not via imitation but robot-robot storytelling, in an effort to model this very human mode of cultural transmission. In this paper we give an account of the methods and inspiration for these experiments, the experiments and their results, and an outline of possible directions for this programme of research. It is our hope that this paper stimulates not only discussion but suggestions for hypotheses to test with the Storybots.
\end{abstract}


\section{Introduction}
In this paper we describe two sets of experiments with small groups of real robots, conducted over the course of more than 10 years, in the Bristol Robotics Lab. The long-term aim of these ongoing experiments is to explore aspects of the question 'how do we have culture?' in a new way, by modelling the low-level processes and mechanisms of cultural evolution, with robots.

The first set of experiments we describe were directly inspired by the thought experiment in \cite[p.106]{Black1999}, which imagines a group of robots capable of imitating each other. Referred to as Copybots, their ability to imitate actions with variation makes them very simple meme machines. Another source of inspiration was Gabriel Tarde who proposed “a remarkable sociological research project” \citep{BarrThrif2007} when he wrote “If we wish to make sociology a truly experimental science and stamp it with the seal of exactness, we must, I believe … write out with the greatest care and in the greatest possible detail the succession of minute transformations in the political or industrial world, or some other sphere of life, … in (our) native town or village, beginning in (our) own immediate surroundings” (quoted in \cite{BarrThrif2007} p.511). Barry and Thrift write that “Tarde’s project was not, as far as we are aware, ever carried out”. When we started to work on what became known as the Artificial Culture project, we realised that we could set up a free-running group of robots (an artificial society) and literally observe, record and analyse every minute detail of the robots’ interactions with each other.

A second and more recent set of experiments extends our robots’ cognitive capabilities with simulation-based internal models. A simulation-based internal model (literally a robot with a simulation of itself, inside itself), allows a robot to be able to ask itself ‘what if’ questions. This capability has been described as a functional imagination \citep{MarqHoll2009}, as it enables a robot to ‘imagine’ the consequences of its actions (and – in our implementation – the reaction of others to those actions). Our experimental implementation of a simulation-based internal model, which we refer to as a consequence engine (CE), has proven to be remarkably powerful. Our experiments with the CE were inspired by both the simulation theory of cognition \citep{HESSLOW2002242,HESSLOW201271} and Dennett’s Tower of Generate-and-Test \citep{Denn1995}. Both simulation and the loop of generate-and-test are present in the architecture of the CE. In our current work, also using the CE, we aim to explore social learning not via imitation but robot-robot storytelling in an effort to model this very human mode of cultural transmission. In addition to the CE our ‘Storybots’ are equipped with what \cite{Penn2008} call the “spectacular scaffolding provided by language”.

Our method for both sets of experiments is to build a working model or, as we prefer to describe it, an \textit{embodied simulation} of a group of autonomous robots, in which the robots are programmed with simple behaviours and interact with each other in an artificial arena. The arena is equipped with a system that allows each robot’s movements to be tracked and recorded, alongside a time-stamped record of each robot’s internal decisions sent to the logging system via a local WiFi network. In this way we are able to capture Tarde’s ‘minute transformations’ for analysis. 

Embodiment is important to us for several reasons. First because experiments with real robots are noisy and unpredictable. Even though our robots are seemingly identical, small differences between the motors and sensors mean that each robot will move and sense in slightly different ways. And the noise will prove to be of critical importance. Unlike computer simulations, the noise, stochasticity and physics come for free, just as they do for animals and us. Second, and perhaps most importantly, robots –- like animals –- have physical bodies that constrain how they behave and ‘think’ \citep{PfeiBon2006}. Robots also see each other only from their own first-person perspectives. Yes, our robots have distinctly nonhuman minds \citep{Penn2008} –- albeit of a kind so simple that the term ‘mind’ is hardly appropriate –- but, we contend, they have enough in common with animals and humans to allow us to plausibly model interesting aspects of social learning and behavioural evolution. The work of this paper fits, we believe, within the microevolution strand of the science of cultural evolution \citep{mesoudi_whiten_laland_2006}, and although simulation models of cultural evolution are not new \citep{Acerbi2020}, we believe that our approach using robots in an embodied individual-based simulation is novel. 

This paper proceeds as follow. In section 2 we outline the Copybots and the key findings from the first set of experiments. Then, in section 3 we describe the Consequence Engine – the key innovation of our second generation of experimental work. We illustrate the CE and the kind of emergent behaviour that is a typical of real-robot embodied simulations with the pedestrian experiment, before then introducing the Storybots. We conclude the paper in 2 parts: in section 4(a) we outline Dennett’s tower of generate-and-test before then showing how it provides a unifying framework in which we can classify all of the robots of this paper. Finally in section 4(b) we discuss experimental possibilities for the Storybots hoping this will stimulate research questions in cultural microevolution that might feasibly be explored.

\section{Copybots}
\label{sec:copybots}

\begin{figure}[t]
	\centering
	\includegraphics[width=0.8\textwidth]{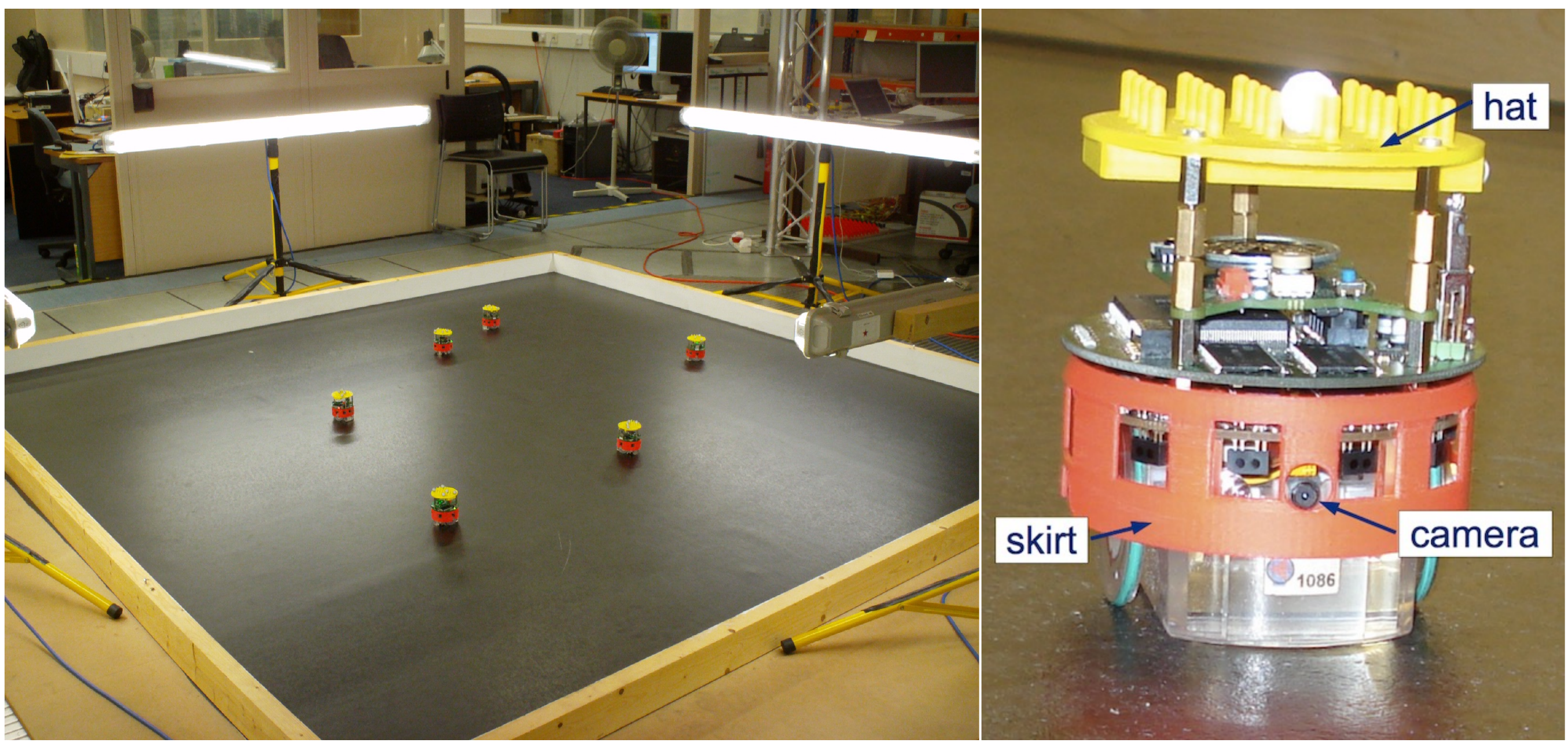}
	\caption{(Left) Artificial culture lab showing 6 robots in the arena. (Right) An e-puck robot, fitted with a red skirt which makes it easier for robots to see each other, and a yellow hat which provides a matrix of pins for the reflective spheres that allow the tracking system to identify and follow each robot.}
	\label{fig:fig1}
\end{figure}

In a series of experiments, we implemented social learning in a group of robots \citep{WINFetal2011}. Simple wheeled robots (see Fig. 1) were programmed to learn socially, from each other, by imitation. These miniature robots –- called e-pucks –- are extremely simple compared with animals; they have just two wheels and so cannot interact with objects in their environment except by colliding with them. Their sensorium is equally limited –- they ‘see’ only with a single 640x480 resolution camera, and something approaching a sense of touch is provided by eight short-range infra-red proximity sensors mounted around their body radius \citep{Mond2009}. From a behavioural perspective the robots are a blank canvas. They have no built-in or innate behaviours, all must be programmed \citep{Winf2012}.

In these experiments imitation was strictly embodied. Robots have no access to each other’s internal states, instead robots observed each other using their onboard sensors and, on the basis of only visual sense data from a robot’s own camera and perspective, the learner robot inferred another robot’s pattern of movements. 
We ‘seed’ each Copybot with initial behaviours, which are self-contained movement sequences (or ‘dances’) which we refer to as memes (\cite{Dawk1976} defines a meme as “that which is imitated”). We then free run the Copybots with each robot alternating between enacting memes and watching (and learning) those memes. 

Not surprisingly embodied robot-robot imitation is imperfect. A combination of factors including the e-puck robots’ low-resolution camera, variations in ambient lighting, heterogeneities among the robots, multiple robots sometimes appearing within a learner robot’s field of view, and of course having to infer another robot’s movements by tracking the relative size and position of that robot in the learner’s field of view, lead to imitation errors. Furthermore, some memes are easier to learn, by imitation, than others (think of how much easier it is to learn the steps of a slow waltz than the tango by watching your dance teacher). The fidelity of embodied imitation for robots, just as for animals, is a complex function of four factors: (1) the behaviours being imitated, (2) the robots’ sensorium and morphology, (3) environmental noise and (4) the inferential learning algorithm. 

But rather than being a problem, noisy imitation was our aim. We are interested in the dynamics of social learning, and in particular the way that memes evolve as they propagate across the collective, by social learning. Noisy social learning means that behaviours are subject to variation as they are copied from one robot to another. Multiple cycles of imitation (robot B socially learns behaviour $m$ from A, then robot C learns the same behaviour $m'$ ($m$ mutated), from robot B, and so on), gives rise to behavioural heredity. And if robots are able to select which learned behaviours to enact we have the three Darwinian operators for evolution, except that this is behavioural, or memetic, evolution. 

\begin{figure}[h]
	\centering
	\includegraphics[width=0.9\textwidth]{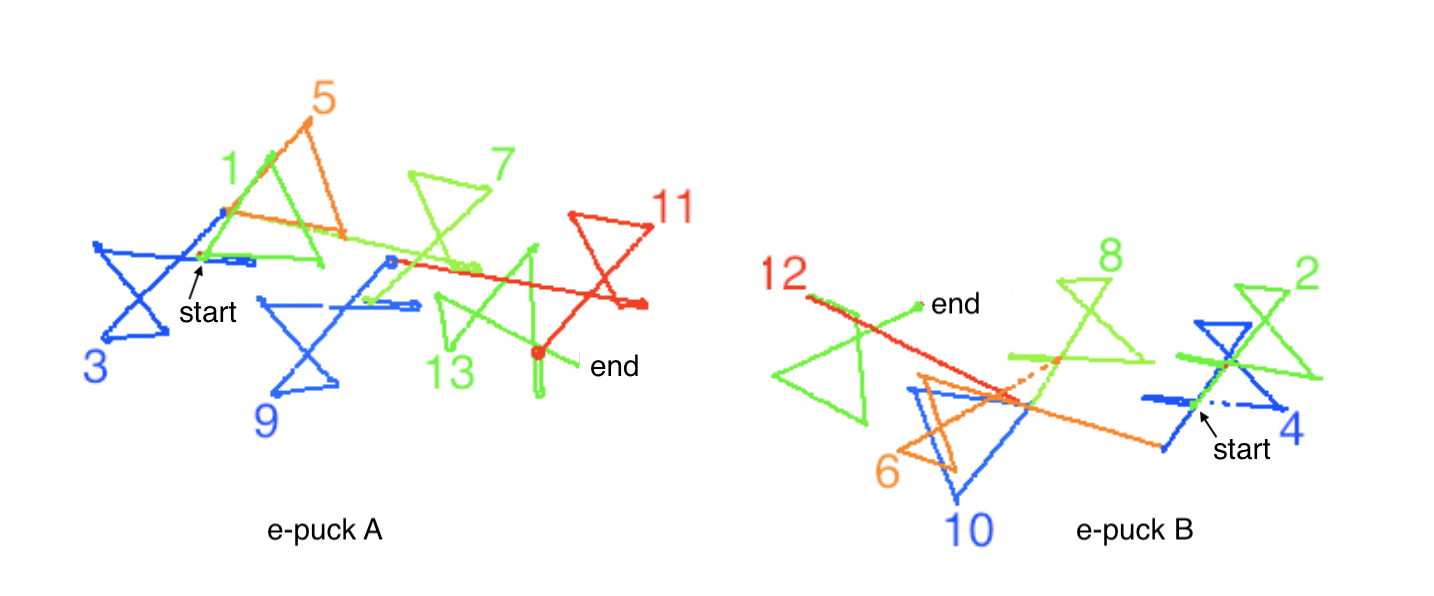}
	\caption{Trajectory plot: two-robot movement-meme evolution in which all copied memes are stored, and meme selection is random with equal probability. The experiment starts with e-puck A (left) in teacher mode and e-puck B in learner mode (right), following a pre-programmed movement trajectory (seed) that describes a triangle (1) with sides of 15 cm. Each meme is shown in a different colour. By chance the triangle meme (1) is imitated imperfectly by e-puck B as an angular figure-of-eight meme (2). This is followed by a high-fidelity imitation (2) -> (3), and again (3) -> (4), so the figure-of-eight meme becomes dominant in the two robots’ collective memory.}
	\label{fig:fig2}
\end{figure}

Our experiments demonstrate that embodied behavioural evolution does indeed take place. If selection is random, that is robots select which behaviour to enact from those already learned – with equal probability – then we see several interesting findings. First, if by chance one or more high fidelity copies follow a poor fidelity imitation, the large variation in the initial noisy learning can lead to a new behavioural species, as shown here in Fig 2. Thus, demonstrating that noisy social learning can play a role in the emergence of new – and potentially useful behaviours in behavioural (i.e., cultural) evolution \citep{WinfErbas2011}. Second, we observe that behaviours adapt to be easier to learn, i.e., better ‘fitted’ to the sensorium and morphology of the robots \citep{ErbasWinf2011}.

\begin{figure}[h]
	\centering
	\includegraphics[width=0.8\textwidth]{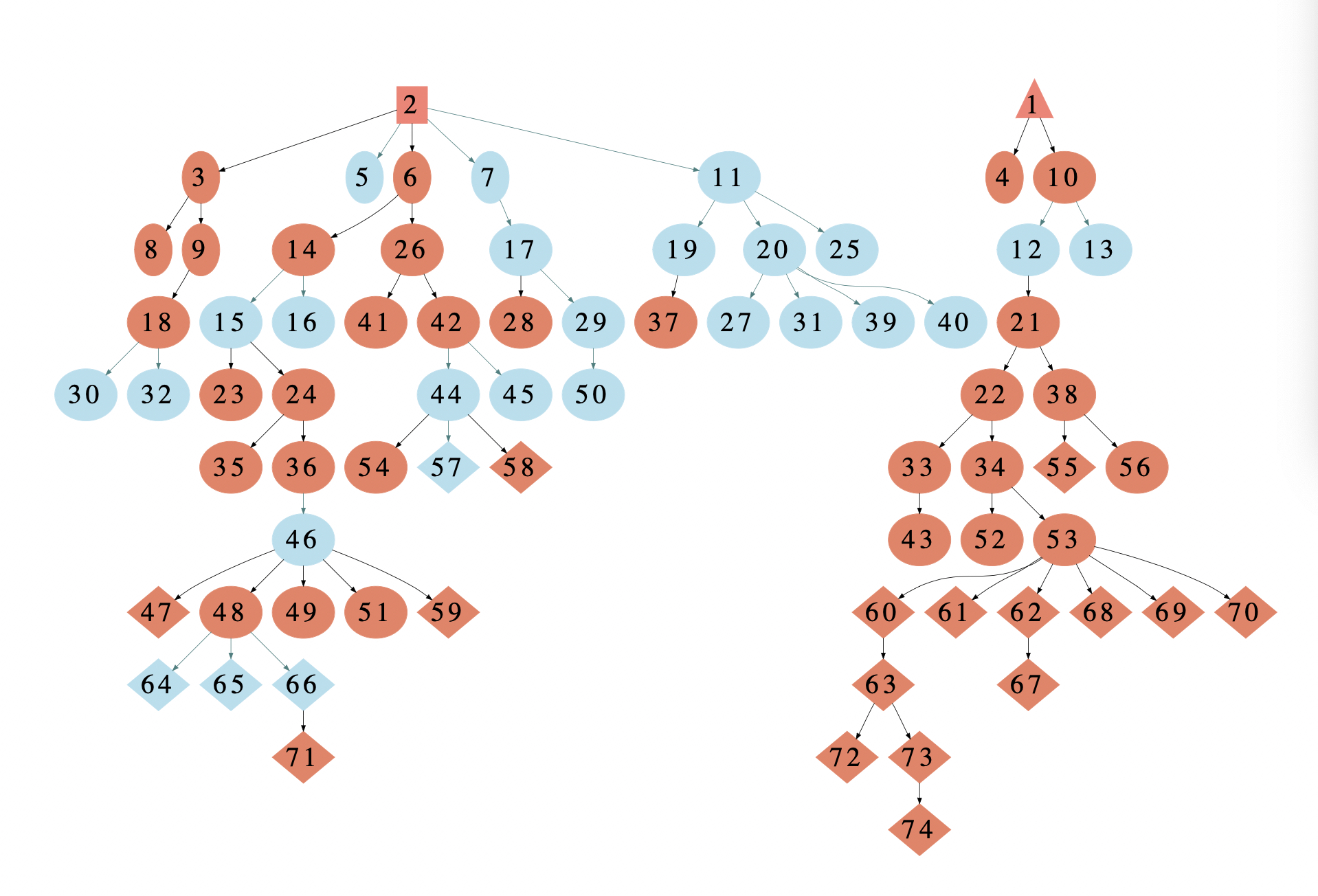}
	\caption{Behavioural evolution map following a 4-robot experiment with limited memory; each robot stores only the most recent 5 learned behaviours. The four robots are each seeded with two behaviours: a triangle dance (labelled 1) and a square dance (labelled 2), from which all memes are descended. Orange nodes are high fidelity copies, light blue nodes are low fidelity copies. The 20 behaviours in the collective memory of all 4 robots at the end of the experiment are highlighted as diamonds. Note one cluster of 12 closely related memes (55), (60-63), (67-70) and (72-74) all descended from (1). Diagram from \cite{Erbas2011}}
	\label{fig:fig3}
\end{figure}

A third finding from this series of experiments is perhaps the most unexpected. When we ran the same embodied behavioural evolution with three memory sizes: no memory, limited memory and unlimited memory, the limited memory case led to the most ‘stable’ population of behaviours across the robot collective, i.e. a smaller number of larger clusters of related memes; in other words, a small number of relatively persistent behavioural types. In Fig. 3 we see one cluster of 12 closely related memes. This result suggests the intriguing conclusion that forgetting may be a valuable trait in behavioural evolution \citep{Erbas2011}. 

A related series of experiments combine social and individual learning. We extended reinforcement learning with imitation, so that robots could observe and learn, by imitation, from more ‘experienced’ individual learners. Reinforcement learning is a well-known approach to machine learning based on trial-and-error interactions between an agent and its environment \citep{Kael1996}. As above, the imitation is strictly embodied, and an imitating robot has no access to the internal state of an observed robot. In a series of experiments, we saw that robots with imitation-enhanced reinforcement learning learned faster than those with reinforcement learning alone. Not a surprising result; social learning is very much faster than individual learning, and robots, just like humans, can benefit from learning socially from more experienced others \citep{Erbas2014}. However, we were surprised to observe that errors in the imitation phase sometimes led to robots learning even faster. It appears that making a mistake while copying another robot can lead to faster learning.

This work has, perhaps for the first time, studied embodied social learning, by imitation, in real-robot collectives. The work has value in extending techniques for robot-robot learning. But its primary purpose is to model and illuminate low-level processes and mechanisms of behavioural evolution. Embodied social learning provides minimal but sufficient biological plausibility and as outlined here, embodiment leads naturally to imperfect imitation, which appears to play an important role in the dynamics of behavioural evolution. 

\section{Storybots}
\label{sec:storybots}

In more recent work we have extended the robots’ cognition with a simulation-based internal model. Robots equipped with a simulation-based internal model have the ability to simulate (or ‘imagine’) the future actions of both themselves and others, and the consequences of those actions.

\begin{figure}[h]
	\centering
	\includegraphics[width=0.8\textwidth]{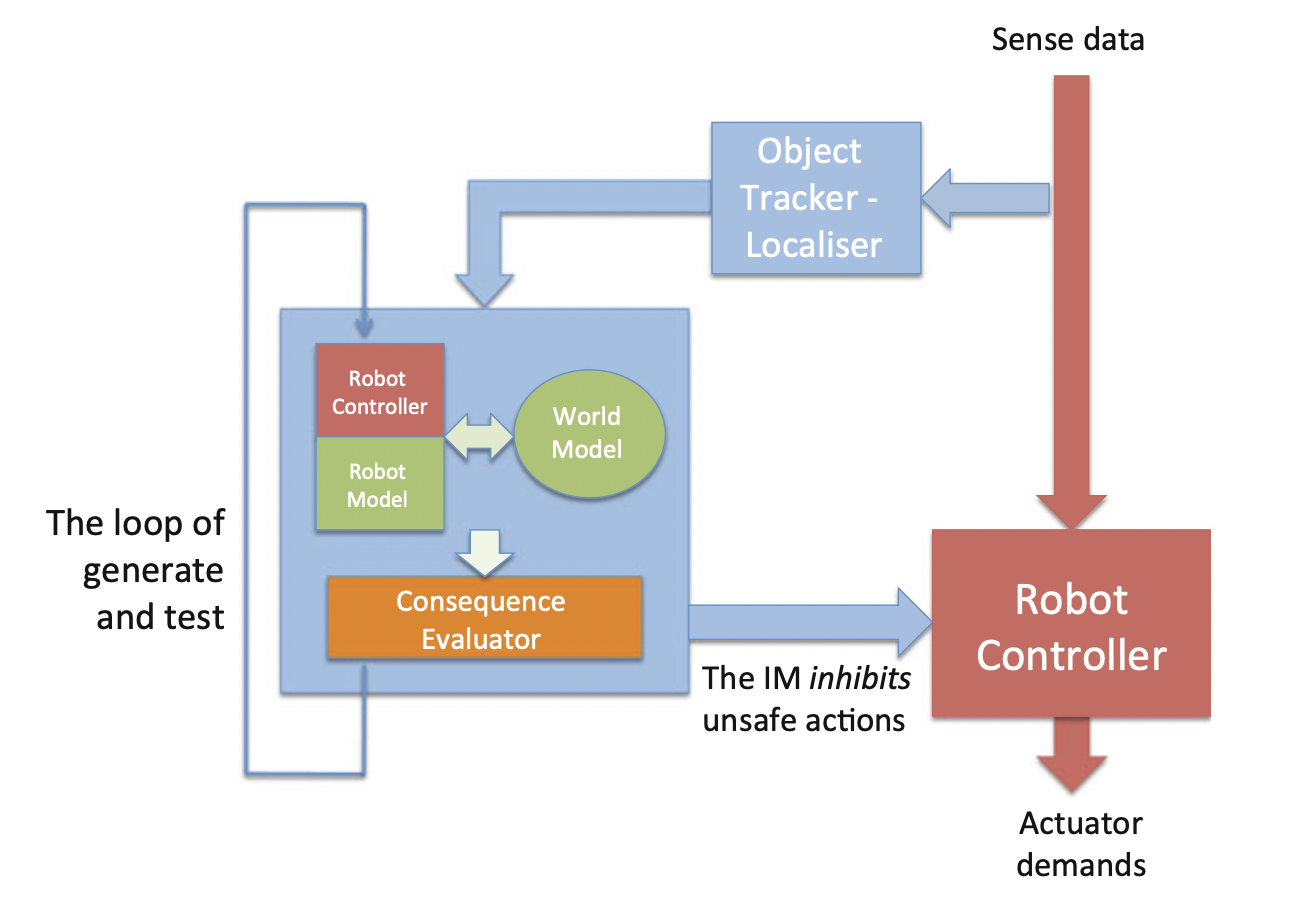}
	\caption{Block diagram of a Robot equipped with a Consequence Engine. The Robot Control dataflows are shown in red (right); the Consequence Engine and its dataflows in blue (left). Diagram from \cite{Win2018stories}.}
	\label{fig:fig3}
\end{figure}

Fig. 4 shows a block diagram of a robot equipped with a CE, which consists of the blue boxes and dataflows on the left. The simulator at the heart of the CE contains three components: a model of the world, which must be initialised to mirror the robot’s immediate environment including the objects and actors in it, as it is now, via the ‘object tracker-localiser’; a model of the robot itself, and an exact copy of the robot’s controller. The loop of generate-and-test shown on the left, generates each of the robot’s next possible actions, then ‘runs’ the simulator for each of those actions in turn. The consequence evaluator determines the anticipated outcome for each of those actions, so that the robot’s action selection can be appropriately moderated (what counts as appropriate depends on whether the CE’s primary purpose is keeping the robot safe, or behaving ethically, etc). In our experiments the CE will typically generate-and-test 30 next possible actions and, for each action, simulate 10 seconds into the future. The complete generate-and-test cycle will be repeated every 0.5 second.

The CE has proven to be a remarkably powerful piece of cognitive machinery. With it we have experimentally demonstrated (i) robots that can make simple ethical decisions in order to pro-actively prevent another robot (acting as a proxy human) from coming to harm \citep{winfield2014towards,vand2018}; (ii) robots with enhanced safety \citep{Blum2018Corridor}, and (iii) robots capable of the imitation of goals \citep{VanWin2017}. We have also argued that the CE provides a robot with a simple artificial theory-of-mind \citep{Winf2018a}. These experiments made use of both e-puck (Fig. 1) and NAO robots (Fig. 6).

\begin{figure}[h]
	\centering
	\includegraphics[width=0.8\textwidth]{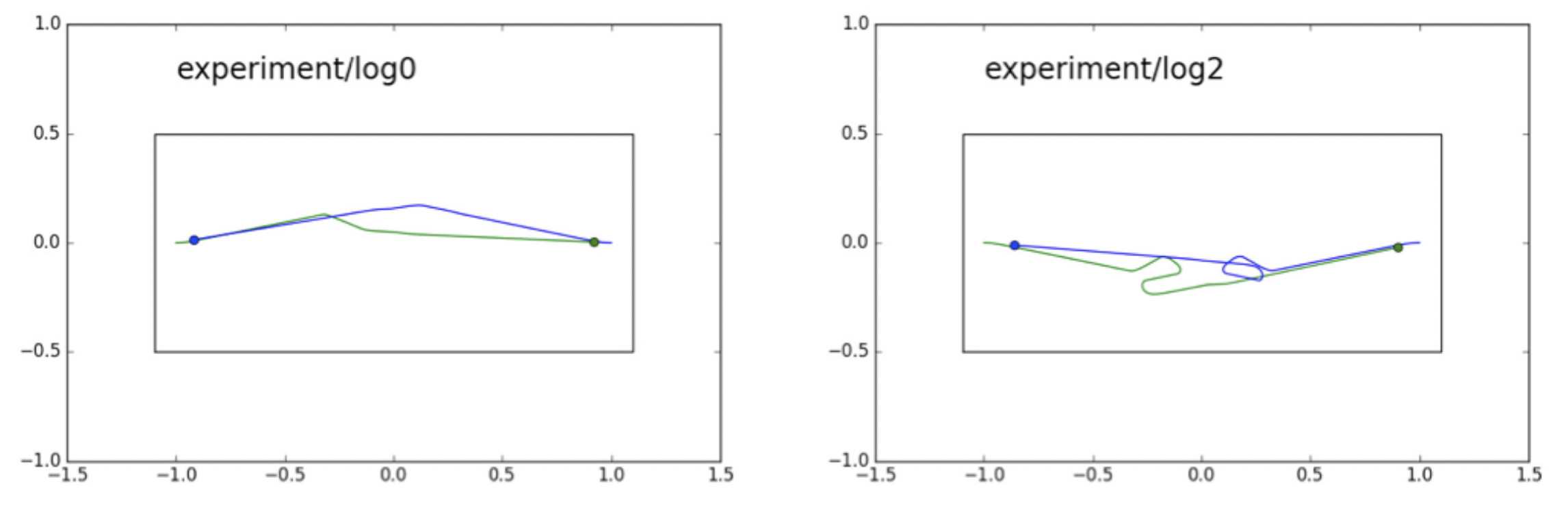}
	\caption{The pedestrian experiment – two trials showing robot trajectories. Two robots, blue and green, are each equipped with a CE. Blue starts from the right, with a goal position on the left, while at the same time green starts from the left with a goal position on the right. (Left) We see the typical behaviour in which the two robots pass each other without difficulty, normally because one robot – anticipating a collision – changes direction first, in this case green. (Right) Here both robots make a decision to turn at the same time, green to its left and blue to its right; a ‘dance’ then ensues before the impasse is resolved. Figure from \cite{Winf2018a}.}
	\label{fig:fig3}
\end{figure}

The Pedestrian experiment provides an elegant example of emergent behaviour in two robots, each equipped with a CE, and programmed with the goal of approaching and then passing each other safely. Fig. 5 shows the trajectories of the two robots: blue, starting from the left, and green, starting from the right. In real-robot experiments, four times out of five blue and green pass each other as two pedestrians would, each stepping to her left (or right), see Fig. 5 left. But one time in five, both blue and green step toward each other, and – just like humans – engage in a short dance before resolving and proceeding on their way (Fig. 5 right). 

More recently we have theorised that the CE may also be co-opted as a mechanism for robot-robot storytelling \citep{Win2018stories}. The CE provides a robot with the cognitive machinery to be able to ask, ‘what if?’ questions. These could be very simple questions like ‘what if I turn left?’. But consider now that instead of the robot using the predictions of the CE to perform (or not perform) some action, it narrates that action together with its imagined outcome to another robot also equipped with a CE, as in ‘If I turned left, I would collide with the wall’. The robot would, in effect, be thinking out loud. Fig. 6 illustrates this process for robot A. 

\begin{figure}[h]
	\centering
	\includegraphics[width=0.8\textwidth]{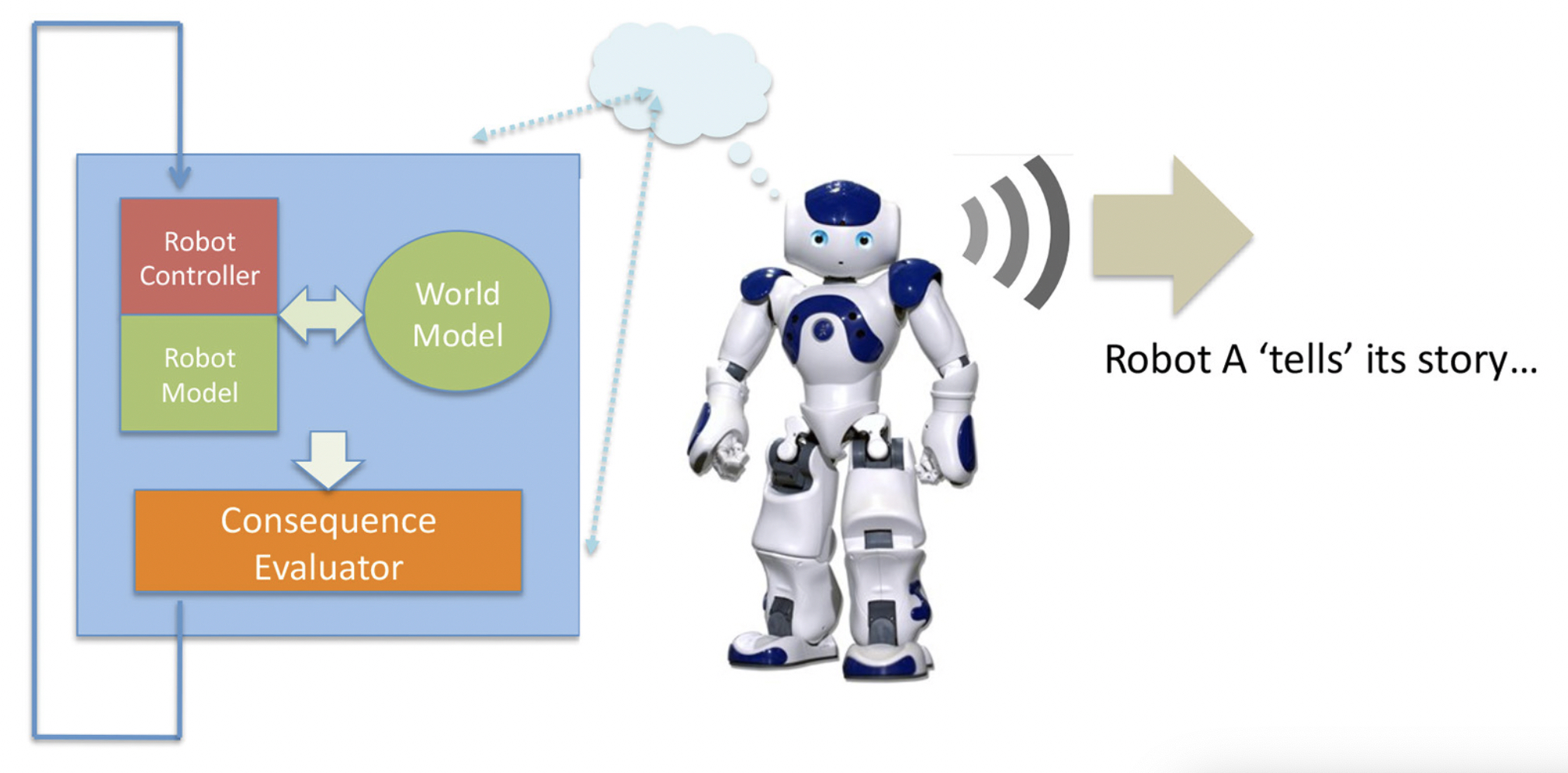}
	\caption{Robot A, the storyteller, ‘narrativises’ one of the ‘what-if’ sequences modelled by its generate-and-test machinery. First an action is tested in the robot’s internal model (left), second, that action—which is not executed for real—is converted into speech and spoken by the robot. Diagram from \cite{Win2018stories}.}
	\label{fig:fig3}
\end{figure}

\begin{figure}[h]
	\centering
	\includegraphics[width=0.8\textwidth]{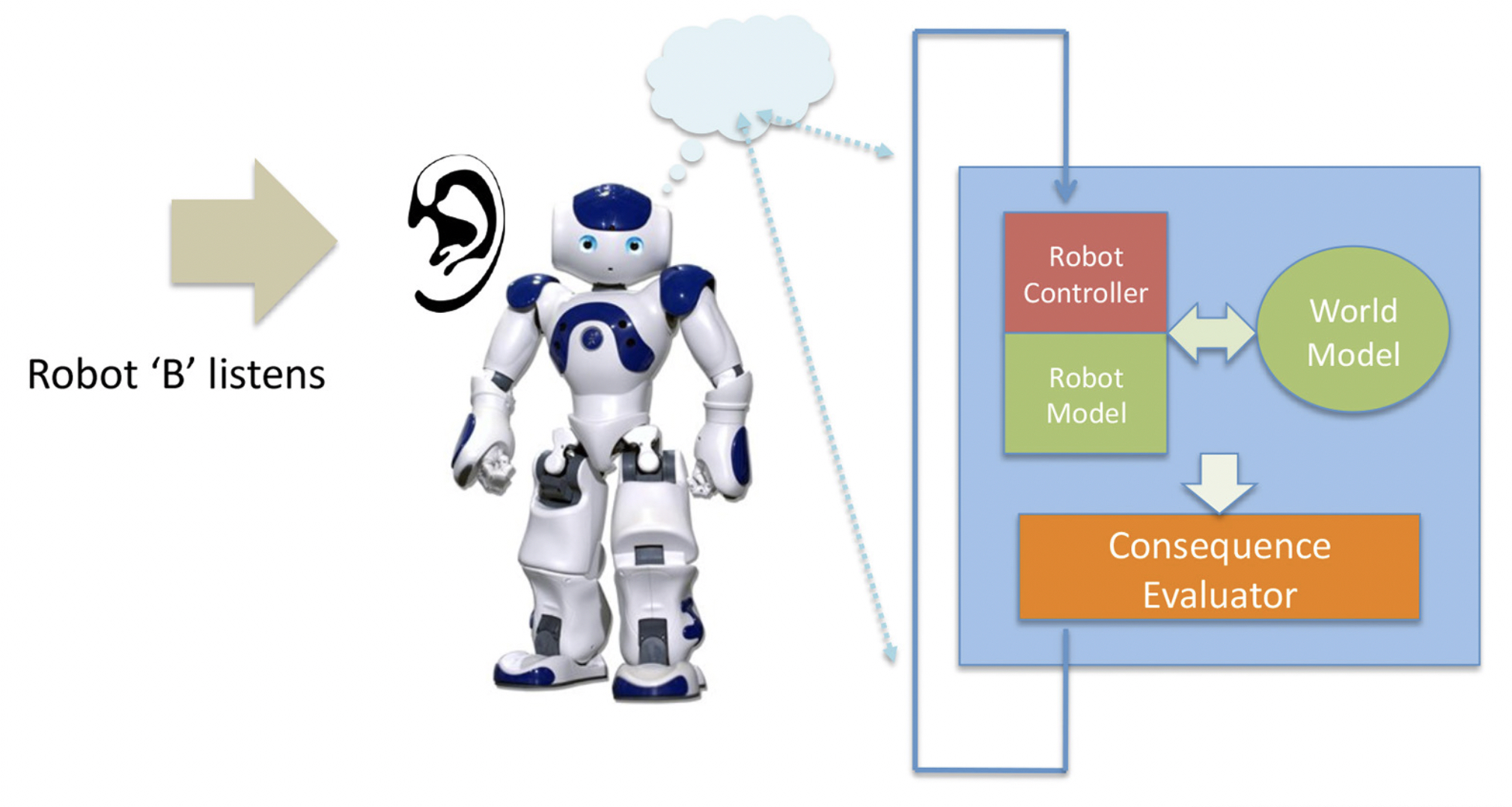}
	\caption{TRobot B, the listener, uses the same ‘what-if’ cognitive machinery to ‘imagine’ robot A’s story. Here the robot hears A’s spoken sequence, then converts it into an action (or series of actions) which is simulated in its own internal model. Diagram from \citep{Win2018stories}.}
	\label{fig:fig3}
\end{figure}

If robot B, the listener, is equipped with a microphone and speech recognition process it is able to listen to robot A’s story, as shown in Fig. 7. Because robot B has the same internal modelling machinery as A – they are conspecifics – it is capable of ‘running’ the story it has just heard within its own internal model. In order that this can happen we need to modify the robot’s programming so that the what-if sequence it has heard and interpreted is substituted for an internally generated what-if sequence. Once that substitution is made, robot B is able to run A’s what-if sequence (its story) in exactly the same way it runs its own internally generated next possible actions, simulating and evaluating the consequences. Robot B is therefore able to ‘imagine’ robot A’s story.  Does this story mean anything to robot B? Arguably it does, as B is able to simulate and therefore ‘experience’ the sensory inputs, and consequences (if any) of listening to A’s story. 

Note that the humanoid NAO robots shown in Figures 6 and 7, do not have human-like intelligence even though their appearance might suggest otherwise. Like the e-puck Copybots, all behaviours must be programmed from scratch. The NAO robots do, however, have the advantage of microphones and loudspeakers, alongside a library of functions for speech recognition and synthesis. This makes them much more suitable as Storybots. 

If we provide not just two, but a group of robots with a rich physical environment they can explore then we are providing the robots with something they can tell each other stories about. And, for the same reasons that our Copybots’ imitation is noisy, so will our Storybots experience imperfect communication, so the stories will mutate as they are told and re-told. The architecture of the CE and its simulation-based internal model opens the possibility that we can replay and visualise any episode in a robot’s ‘imagination’, thus adding further detail to Tarde’s ‘minute transformations’ and allowing us to inspect the robots’ mental representation of stories as they pass from robot to robot.

\section{Discussion and conclusions}
\label{sec:Disc}

\subsection*{(a)	Dennett’s Tower of Generate-and-Test: a unifying framework}

As mentioned in the introduction, Dennett’s Tower of Generate-and-Test directly inspired the second series of experiments outlined above, culminating in the Storybots. Dennett’s tower also provides us post facto with a single framework to unify all of the experimental work outlined in this paper. 

\cite{Denn1995} proposes a conceptual framework, the Tower of Generate-and-Test, for thinking about design options for brains. Each floor of the tower uses the three Darwinian operators: copy, generate variations, test outcomes – repeat. Each floor builds on the outcome of the previous ones. The framework provides a way of seeing how humans, as a cultural species, emerged from creatures with no cumulative culture, using the same ‘generate and test’ process all the way up. The ground floor is inhabited by Darwinian creatures. Variation is provided by the more or less random recombination and mutation of genes, and selection is brutal – design-by-death \citep{Black2010}. All living things are Darwinian creatures. 

Some of these creatures emerged with conditionable plasticity; that is, not all their behaviour was genetically determined. Occupying the first floor of Dennett’s Tower, these Skinnerian creatures try out a variety of responses to their environment selecting only actions that are reinforced for repeating. Dennett named them after Skinner’s comment that, “Where inherited behavior leaves off, the inherited modifiability of the process of conditioning takes over” \citep[p. 83]{Skin1953}. Most plants and animals are Skinnerian, as well as Darwinian, creatures.

On the third floor we find Popperian creatures. These are able to think through possible actions and their likely consequences before carrying them out. This requires internal models of (relevant features of) the environment, as well as of their own body and behaviour. Variation comes from imagining different actions; selection is by imagined consequences. As Popper remarked, when imagining outcomes we “Let our conjectures, our theories, die in our stead” \citep{Popp1977}. We are not alone in being Popperian creatures; most mammals, birds, fish and reptiles can learn through both classical and operant conditioning, and can contemplate the consequences of at least some of their actions.

On the fourth floor are Gregorian creatures. As far as we know, we are the only Gregorian creatures, at least on this planet. Gregory introduced the idea of ‘tools of Mind’ or ‘mind-tools’ by which he meant “aids to measuring, calculating and thinking” \citep[p.48]{Greg1981}, including tools like scissors or levers as well as spoken and written words, and ways of counting. Language makes possible long trains of thought, the ability to look ahead, and the sharing of tools that enhance intelligence. These tools are built up over generations by creatures that can copy information from each other, building up culture. Gregorian creatures are therefore meme machines \citep{Black1999} as well as Darwinian, Skinnerian and Popperian creatures. This ability to imitate and learn from others makes possible what Dennett calls the “deliberate, foresightful generate-and-test known as science” \citep[p.380]{Denn1995}.

Let us now place our robots within the floors of Dennett’s tower. The basic Copybots are Darwinian creatures alone, their design having been selected from many other possible designs we might have chosen – a form of design-by-death. The Copybots with imitation-enhanced learning are also Skinnerian creatures. Both types can imitate, which might suggest they are Gregorian, but this would be a misclassification as the Copybots have no internal model – the defining characteristic of Gregorian creatures. By contrast, all of our robots with a Consequence Engine (CE) are Popperian; the CE enables them to generate and test hypotheses about what to do next. They are, however, not strictly Skinnerian because we have not added reinforcement learning (although this is perfectly feasible). The Storybots proposed in Figures 6 and 7 finally take us from the Popperian to Gregorian level. 

While the Copybots imitate by visually observing movement memes, lacking a CE they cannot predict and evaluate the consequences of those imitated behaviours. The Storybots, on the other hand do not imitate behaviour directly. Their method of learning from others is mediated by the mind-tool of language. Table 1 summarises the classification outlined here. 

\begin{table}[h]
	\centering
	\begin{tabular}{p{8cm}cccc}
		\toprule
		robot     & Darwinian & Skinnerian & Popperian & Gregorian \\
		\midrule
		Basic Copybots (\cite{WinfErbas2011}) & \ding{51} & & &     \\
		Copybots with Imitation-enhanced reinforcement learning (\cite{Erbas2014}) & \ding{51} & \ding{51} & &      \\
		Robots with Consequence Engine (CE) (\cite{Blum2018Corridor,VanWin2017}) & \ding{51} & & \ding{51} &  \\
		Storybots (\cite{Win2018stories}) & \ding{51} & & \ding{51} & \ding{51} \\
		\bottomrule
	\end{tabular}
	\caption{Classifying the robots of this paper within Dennett’s conceptual Tower of Generate-and-Test}
	\label{tab:table}
\end{table}

Applying the framework of Dennett’s Tower does raise the interesting question of whether intelligence must necessarily be achieved by building each level of generate-and-test on top of the preceding one, or whether robot intelligence might skip one or more, to achieve cumulative culture more directly.  

\subsection*{(b) What can we learn from the Storybots?}

In order to address this question, consider first what we might learn from the Storybots as presented in Section 3. These robots are equipped with a CE plus the speech synthesis and recognition capabilities shown in Figs. 6 and 7. Like the other robots with a CE outlined at the start of Section 3, the basic Storybots have a short-term memory which is used to retain the evaluated consequences of each generated-and-tested action, in order that the most appropriate action can be selected. But at the end of each complete cycle of generate-and-test that short-term memory is cleared ready for the next cycle. These robots do not learn, which may seem surprising given the capabilities demonstrated by the CE.  However, even these basic Storybots (like the first Copybots we tested) can usefully allow us to explore how stories vary as they are told and re-told by several robots. As part of this experiment, we would first need to find the optimal balance between zero and perfect fidelity robot-robot speech-transmission (as we had to do with the Copybots for vision-based imitation) such that we do see a reasonable level of variation; this would require for instance adjusting speech loudness, microphone sensitivities and attending to directionality so that the listener robot is facing the narrating robot.

Next consider memory. It would be straightforward to equip each Storybot with a long-term (autobiographical) memory. The memory would store events (things that happened to the robot) and the actions (of the robot). Those actions might be either reactions to events, or actions initiated by the robot in order to achieve some goals. Alongside these actions and events, the robot could also remember stories; not only stories it has heard (with the name of the storytelling robot), but also ‘what if’ stories that the robot has asked itself\footnote{Which would provide an interesting model of remembering by re-imagining \citep{Rath2011}}. Of course, since the robot uses its CE to choose its next possible action then the ‘what if’ stories are closely linked to the robot’s actions. By remembering the ‘what if’ sequence that led to a particular action a robot is, in effect, recalling the \textit{reason} it chose that particular action. Although less useful we could also store the ‘what if’ actions that were tested but not selected.

When it comes to deciding how to select \textit{which} of a robot’s stories it should narrate, we have several options. (i) we could use the same strategy as the Copybots and choose one from the robot’s memory at random with equal probability. If we also limit the Storybots memory, as suggested by the Copybots experiments of \cite{Erbas2011}, then we might expect that – in time – some stories become dominant in the collective memory of the group of Storybots, for no other reason than they happen to have been selected then re-told with high fidelity. Of more interest would be (ii) selecting stories by content, for instance those that point out hazards, so that telling (for the first time) or re-telling the story is, in effect, ‘spreading the word’. The robot would run each of the stories it has heard and remembered, in its CE, and select the one that it ‘imagines’ as the most dangerous, using exactly the same evaluation mechanism the robot would use when (generating and) testing its own possible actions. A third interesting option (iii) would be to select stories to re-tell on the basis of which other robot told the story. One strategy would be to choose to re-tell stories told by the robot whose stories have been re-told the most often in the group, thus introducing a frequency bias. Another would be to re-tell stories from the robot whose stories are judged the most impactful in the sense outlined in strategy (ii). If we introduce new robots into the group at different times we might see the emergence of an ‘elder’ storyteller robot that is accorded a prestige bias. There are many interesting options to be explored within strategy (iii).

Also important is how a robot decides \textit{when} to tell the story it has selected for re-telling. Since our Storybots will, like the Copybots, be moving around in their shared environment they will encounter each other quite frequently. These encounters present opportunities for Storybots to tell new stories or re-tell previously heard stories. How, on meeting each other, would robots agree which one will be the storyteller (as in Fig. 6) and which the listener (Fig. 7)? A simple mechanism would be for both robots – on meeting – to start an internal timer, choosing at random, the number of seconds to wait. If a robot has not heard the other one speak before its timer runs out then it will speak first, taking the role of storyteller. The other robot will hear the storyteller start to speak while its timer is still running and adopt the role of listener. If the experiment uses selection strategy (iii) then a robot could, on encountering a ‘prestigious’ storyteller, add a few extra seconds to its randomly chosen wait-before-speaking timer. We could call this a ‘deference’ value.

By integrating an autobiographical memory within the CE of the Storybots, we are in effect providing the robots with what \cite{Conw2005} calls a self-memory system. Arguably our Storybots will have sufficient cognitive machinery for the emergence of an artificial ‘narrative self’\footnote{Noting that \cite{Conw2005} uses the term ‘Working self’.} that will become, in a short time, unique to each robot. The directions we have outlined here suggest the exciting possibility that we can, with our embodied simulation, experimentally explore the relationship between the developing ‘narrative selves’ of the individual robots and their evolving shared narrative, i.e., oral culture. What might such an experiment tell us about animal or human cultural evolution?

\section*{Acknowledgments}

The work outlined in Section 2 was funded by EPSRC grant ref EP/E062083/1. The authors gratefully acknowledge project co-investigators James Bown, Robin Durie, Frances Griffiths, Jenny Tennant Jackson and Alistair Sutcliffe, and are especially grateful to Mehmet Erbas who conducted most of the experimental work. The work outlined in Section 3 was in part funded by EPSRC grant ref EP/L024861/1. The CE was implemented by Christian Blum, Wenguo Liu, and Dieter Vanderelst, and the pedestrian experiment was coded and run by Mathias Schmerling and Chen Yang. We are also grateful for conversations with Daniel Dennett and the late Richard Gregory who directly inspired the Storybots when he declared “when your robots start telling each other stories \textit{then} you’ll really be onto something”.

\bibliographystyle{unsrtnat}
\bibliography{references}  






\end{document}